# Saliency Based Fire Detection Using Texture and Color Features

Maedeh Jamali, Nader Karimi, Shadrokh Samavi
*Department of Electrical and Computer Engineering, Isfahan University of Technology*
Isfahan, 84156-83111 Iran

*Abstract*— **Due to industry deployment and extension of urban areas, early warning systems have an essential role in giving emergency. Fire is an event that can rapidly spread and cause injury, death, and damage. Early detection of fire could significantly reduce these injuries. Video-based fire detection is a low cost and fast method in comparison with conventional fire detectors. Most available fire detection methods have a high false-positive rate and low accuracy. In this paper, we increase accuracy by using spatial and temporal features. Captured video sequences are divided into Spatio-temporal blocks. Then a saliency map and combination of color and texture features are used for detecting fire regions. We use the HSV color model as a spatial feature and LBP-TOP for temporal processing of fire texture. Fire detection tests on publicly available datasets have shown the accuracy and robustness of the algorithm.**

## I. Introduction

Fire detectors for surveillance systems in indoor and outdoor monitoring have become common in recent years. Conventional detectors usually distinguish the presence of fire particles using photometry or ionization. Typically, these methods are suitable for indoor applications, but they are very sensitive to any small fire or smoke, and the number of false alarm can be high. One of the crucial disadvantages of point detectors is that they are distance-limited. This means that they only alarm when the fire or smoke stimulates them. Therefore, we cannot have real-time detection. As a result of camera technology development and improvement in memory capacity and computing power, video-based systems instead of conventional fire detection methods become a hot research topic in the field of fire recognition. By using video-based methods, we can distinguish uncontrolled fires at an early stage before they turn into disasters. Extracted features from video frames could be sent to a center for further analysis.

Many methods exist for fire detection, which all of them try to improve the accuracy by using different feature vectors. Some algorithms only use color information for distinguishing fire regions [1-5]. Using color information alone is not accurate due to the false detection of fire-colored objects.

Because of the motion characteristic of fire regions, some researchers use motion detection algorithms to improve the accuracy and reduce false alarms of stationary fire-colored objects [6-10]. Some methods try to detect fire and smoke in videos [11-16]. They believe smoke detection can be suitable for early fire alarms because smoke appears before flames. Some other methods have worked in fuzzy space for fire detection [17-18]. Because of irregularity and working in continues spaces, using a fuzzy approach can be a reasonable way. Most of these methods used motion detection algorithms; hence they cannot be used for moving cameras. Proposed methods of [19-22] are based on temporal and spatial information for fire detection. Reference [19] used a covariance matrix-based method in 3D blocks and used this for stationary and non-stationary cameras. Recently, researchers try to use CNN based methods in fire detection [23-25].

In this paper, we use a new method for detecting fire regions. We use a saliency detection algorithm for separating probable areas that are supposed to contain a fire. After separating probable fire regions as salient parts, we use HSV color space to distinguish fire regions based on color features. Also, by using a superpixel algorithm, we improve those regions, which had some missing parts in previous phases. In the last stage, we use a texture-based algorithm, named local binary pattern three orthogonal planes (LBP-TOP), as a filter which determines the presence of fire texture in a 3D block.

The rest of the paper is organized as follows: In section II we present our proposed method. Each step of our approach is explained in this section by details. Section III provides experimental results and comparisons with other fire detection algorithms. We also offered a few suggestions for possible future works in section IV. The paper is concluded in section V.

## II. PROPOSED METHOD

In this section, we introduce our proposed method. First, we briefly explain a fast saliency detection algorithm that is used for separating conspicuous areas in each frame. We use this phase instead of motion detection methods. The advantage of using this phase is that it can also be used with non-stationary cameras. In subsection B our color model is described, and subsection C is dedicated to LBP-TOP texture classification, which is trained by SVM. Finally, we present our feature vector for classifying fire regions. Figure 1 shows the main steps of our algorithm.

### A. Saliency Detection algorithm for separating fire regions

The main idea in saliency detection algorithms is to identify the most significant and informative parts of a scene which indicate human fixation locations over an image. These methods are guaranteed to give compact and homogeneous objects over an image [26-28]. Saliency detection methods can help us to extract probable fire regions in each video frame. The reason for using saliency detection is to distinguish fire areas by using the contrast that exists in fire regions with respect to the surrounding parts. By using this method, we can extract connected and homogeneous regions. Salient objects extraction is done based on the content of the entire image without any learning. So we can remove some off-line learning which is used for motion detection in many fire detection methods. By doing so, we can use this method for other applications where non-stationary cameras are used.

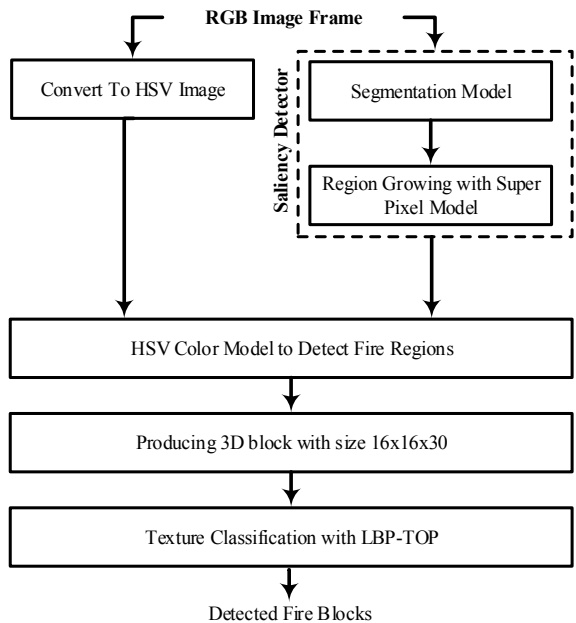

Figure 1. Block diagram of the proposed method.

We tested different saliency methods which were presented their results here. Figure 2 shows the results of the methods of [26-28]. As can be seen in Fig. 2, the output of [26] is better than the other methods. It is capable to detect salient regions that are not in the center of the image.

All of these methods produce a saliency map in grayscale format whereas we need a binary mask for further analysis. Hence, we propose the following model for more precise extraction of salient fire regions. This model contains two steps which are explained in the following.

*1) segmentation model*

We need a suitable mask for segmentation purposes. Based on experimental results, we found out that it is better to use local and global information for segmentation. As Fig. 2 shows, fire regions are the brightest regions in the saliency map. Therefore, we need a method that can extract these regions. If only global information is used, some undesirable areas are classified as fire regions. We also use local information to obtain better segmentation. For this aim, each frame is divided into 32×32 pixels blocks (all of the frames are 320×240 pixels). After partitioning of the frame, each block is analyzed and the mean and variance of them are calculated. Since we want to extract the brightest regions, we choose the highest mean among all blocks. But we expect that some surrounding blocks are also members of the fire regions. We, therefore, use a global threshold on the saliency map. Otsu threshold is used for the global thresholding. The action of this segmentation is summarized in the following equations:

$$Max_{mean} = max\{local\_mean_1^i\} \quad i = 1 \text{ to number of blocks} \quad (1)$$

$$fire_{mask} = \begin{cases} 1 & saliencyMap \geq Max_{mean}\text{-}GT \\ 0 & o.w. \end{cases} \quad (2)$$

which *saliencyMap* is a greyscale map that is produced based on [26], and *GT* is the global threshold based on Otsu measurement on the input frame. In the following steps, we only process those regions in $fire_{mask}$ that are marked as 1. Figure 3 shows the output of our segmentation.

*2) Region growing using superpixel extraction*

After extracting fire regions by saliency detection, a region growing method is used to complete the identification process. For this aim, we use the quick-shift algorithm [29]. This algorithm is a segmentation method for finding superpixels. Occasionally we observe that after segmentation, some parts of fire regions are not considered. We propose to use a region growing method for adding the omitted sectors. The quick-shift algorithm can extract superpixels in an image. This algorithm has several parameters that we modify them to work best for our application.

After the saliency map is obtained, the quick-shift is applied to superpixels. Then we check superpixel overlaps with the salient region. If a superpixel has more than 33 percent overlap with the salient region then the saliency region in the mask is corrected to cover the whole of that superpixel. This procedure is repeated for all superpixels. Part (d) of Fig. 3 shows the final mask after applying the quick-shift procedure. After extracting probable fire regions, a color model is used to filter out the mistakes which are produced in the first stage. In the following, we explain the color model.

*B. HSV color model*

Fire pixels have specific color characteristics that can help to identify fire regions more precisely. Fire regions are usually brighter than other areas in the image. Therefore we need a color space to better model this feature. HSV (hue-saturation-value) color space is a suitable candidate for this purpose because it separates the brightness information from color information.

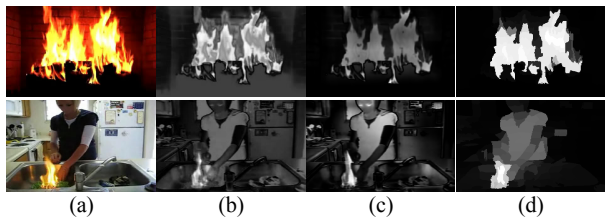

(a)      (b)      (c)      (d)

Figure 2. The output of different saliency algorithms for fire regions extraction, (a) Original image, (b) output of saliency method in [27], (c) output of saliency method in [28],(d) output of saliency method in [26].

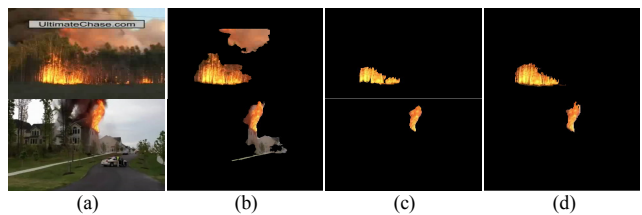

(a)      (b)      (c)      (d)

Figure 3. The output of proposed segmentation, (a) original image, (b) segmentation method using saliency map in [26], (c) first segmentation, (d) the output of proposed method with region growing using quick shift algorithm.

To find the characteristics of fire regions, we use nearly a set of 170000 fire pixels in 500 frames. First, we manually select the fire regions in each frame and convert them to HSV color space. Then a histogram of fire pixels is created for each of the three channels. After normalization of color values in different channels, we find out the range of values in fire regions. The hue of fire pixels usually is between 0 and 0.2, and the two other channels approximately have values between 0.6 and 1. We use Otsu thresholding to extract the fire region. The following formulas show how thresholds are set.

$$hue_{fire} = \begin{cases} 1 & 0 \leq I_{hue} \leq gt_{hue} \\ 0 & o.w. \end{cases} \quad (3)$$

$$saturation_{fire} = \begin{cases} 1 & gt_{saturation} \leq I_{saturation} \leq 1 \\ 0 & o.w. \end{cases} \quad (4)$$

$$value_{fire} = \begin{cases} 1 & gt_{value} \leq I_{value} \leq 1 \\ 0 & o.w. \end{cases} \quad (5)$$

which $gt$ is a global threshold that is based on Otsu method and for each channel is calculated separately. This parameter depends on an image and has different values for different images. Parameter $I$ indicates the input image, and its index shows the respective channel.

Those pixels that satisfy all of these conditions are considered as flame pixels. The combination of (3), (4) and (5) are used to extract fire-color regions. Figure 4 shows the output of this step. This step also gives us a binary mask that we combine with the salient mask to detect probable fire regions. Logical AND operator is used for this combination.

### C. Fire texture classification using LBP-TOP

After applying two previous phases, the fire regions and some fire-color areas are extracted and we are not losing any fire areas. But we need to discard fire-colored objects which are not a real fire, which should reduce the false positive rate. For this aim, the texture of the fire is considered. Our experimental observations show that fire areas have a special texture, which makes them different than non-fire regions with the same color.

We both consider temporal characteristics of texture as well as the appearance of the texture. Local Binary Pattern (LBP) is a descriptor that is used for pattern extraction in images. But we need a method which can extract patterns in time. Local Binary Pattern from Three Orthogonal Planes (LBP-TOP) is a descriptor with such ability that is used for fire texture extraction in 3D blocks. The LBP-TOP method is described in detail in [30-31]. For our application, we partition videos into Spatio-temporal patches. In each 3D block, LBP-TOP is extracted and 8 pixels are considered as neighbors. $xy$ plane contains appearance features, whereas $xt$ and $yt$ contain motion characteristics with limited appearance features. All of our fire videos are scaled to $320 \times 240$ so that each frame is divided into 300 blocks (each block being $16 \times 16$ pixels). We use 30 consecutive frames for temporal analysis of fire regions. Our 3D blocks do not have any overlap in the spatial domain, but in the time domain, they have overlap with 15 frames. Figure 5 shows the process of using LBP-TOP for fire texture extraction in a video sequence.

As shown in Fig. 5, three orthogonal planes are used to extract fire textures in the frontal view and time domain. In the first stage, after applying two pre-processing phases, we divide videos to $16 \times 16 \times 30$ blocks, and then the LBP-TOP algorithm is used to extract fire texture. Three LBP histograms from $xy$, $xt$, and $yt$ planes are obtained. Then these histograms are concatenated to form a feature vector. Now we have 300 blocks but some of them do not contain any useful information. Therefore we do not need to consider all of the blocks. We only consider those 3D blocks where more than 1/8 of its pixels are salient. By using this method, we do not need to process all of the blocks which makes the algorithm more fitted for real-time processing.

A classification method is required to label each 3D block as a fire or non-fire block. In our proposed method we use SVM classification for this purpose. We use three different kernel functions for the SVM classification and compare their performance in the classification of fire blocks. We use 5-fold cross-validation on 3500 blocks, which are selected manually and have two labels of "1", for fire block, and "−1" for non-fire blocks. TABLE I shows the comparison between linear, polynomial, and RBF kernel functions. We used $2^{nd}$ and $3^{rd}$ degree polynomials and RBF with radii of 5, 10, and 15.

Based on TABLE I, we see that accuracies of polynomial and RBF functions are more than the linear kernel. Polynomial kernel of $2^{nd}$ degree produces better results in different criteria and is simpler and faster in comparison with the RBF kernel with radius 10. Hence this kernel is selected for SVM classification.

After using LBP-TOP each block is labeled as a fire block or non-fire block. If we find a fire block in this phase, we conclude that the 3D block contains a fire. Therefore the fire regions that were extracted using the saliency map and color model are valid fire regions. Fire cannot appear in one frame and suddenly disappear in the next frame. Therefore, if we do not detect fire in 15 frames but previous frames contained a fire, we label these frames as fire and refine the decision of the system. This means that our decision is based on the history of previous detections.

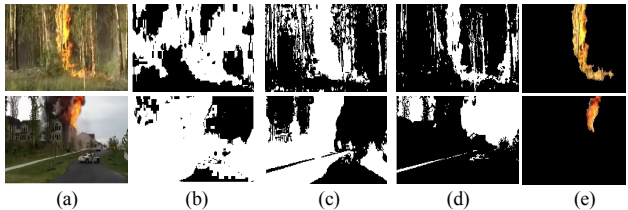

Figure 4. Output of color model, (a) original image, (b) output of hue mask, (c) output of saturation mask, (d) output of value mask, (e) combination of three masks.

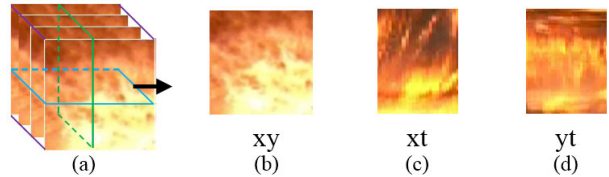

Figure 5. LBP-TOP computation in fire videos, (a) video frames and three orthogonal planes, (b) image in xy plane, (c) image in yt plane, (d) image in xt plane.

Table I. Comparison Between Different Kernel Function

| Kernel function | linear | polynomial | | RBF | | |
|---|---|---|---|---|---|---|
| | | Degree of polynomial | | Sigma value | | |
| | | 2 | 3 | 5 | 10 | 15 |
| TP | 6267 | 6619 | 6584 | 6863 | 6694 | 6501 |
| TN | 6554 | 6746 | 6776 | 5747 | 6636 | 6631 |
| FP | 446 | 381 | 224 | 1253 | 364 | 369 |
| FN | 733 | 254 | 416 | 137 | 306 | 499 |
| Sensitivity | 0.895 | **0.946** | 0.941 | 0.980 | 0.956 | 0.929 |
| Specificity | 0.936 | **0.964** | 0.968 | 0.821 | 0.948 | 0.947 |
| Accuracy | 0.916 | **0.955** | 0.954 | 0.901 | 0.952 | 0.938 |
| Error | 0.084 | **0.045** | 0.046 | 0.099 | 0.048 | 0.062 |

III. EXPERIMENTAL RESULT

The proposed method is implemented in MATLAB R2016a and is tested on videos in the Firesense project. Also, we used additional videos that show different conditions and are recorded with a moving camera. These video sequences contain a variety of environmental conditions, such as day time and night time. This dataset includes a variety of videos with different situations that can evaluate the performance of a fire detection method. Appropriate block size is required in the third phase of the algorithm. Some small fire regions cannot be revealed in large blocks. On the other hand, if the very small block size is chosen, it can be time-consuming. Therefore, we tested blocks with two different sizes of $8 \times 8 \times 30$ and $16 \times 16 \times 30$. We used 200 fire blocks and 200 non-fire blocks with size $16 \times 16 \times 30$, which are manually selected. Each block is classified using our method, and we find out that the block size of $16 \times 16 \times 30$ produces better results.

In a fire detection method, an algorithm has a good performance when its detection rate is high. Our algorithm achieves an average sensitivity or detection rate of 98.5% using the test video samples, which shows the high sensitivity of our algorithm.

In this section, we compare our method with some recent methods such as [2] and [19-22]. These methods also used Spatio-temporal features. In our method, if LBP-TOP detects fire in a 3D block, we conclude that the fire exits in the frame and the fire regions are detected using a combination of saliency and HSV color model. Figure 6 and Fig. 7 shows a comparison between our results and those of other methods. For non-fire videos, lower values show better results that mean lower false detection. It can be seen that our approach has a higher performance in different scenarios. Table II shows the comparison of our method with [22] and [2] based on the True Positive and True Negative rate due to fire and non-fire videos, respectively. Some of these videos are not reported in these papers, so we do not have values for these videos. As we can see, our method has better results based on reported values.

Fig. 8 shows the output of our algorithm on two fire and non-fire videos. Images in column b of Fig. 8 show the output of LBP-TOP, which determines blocks contain flames in the previous 30 frames. If any white block exists in these images it means that fire exists in these frames and the flame is extracted by color and saliency model; otherwise, these frames do not have flames.

IV. FUTURE WORK

For future works, we can implement the flame detection algorithm in wireless sensor networks [32]. In a camera sensor network, we can consider the compression of captured flame images [33]. We can also investigate on routing algorithms for fire detection using camera sensor networks [34]. One could also use the proposed color features of this paper for image segmentation applications [35].

V. CONCLUSION

In this paper, a video fire detection system was proposed based on the detection of saliency and extraction of texture by LBP-TOP. It was shown that the use of saliency detection was a good choice for extracting probable fire regions. This was because of the high contrast that exists between fire and its background. We also showed that saliency detection worked based on an individual frame without any prior knowledge or any need for a learning phase. The color model was also an essential part of our proposed method, which eliminated non-fire regions from the decision process. An essential aspect of this research was the combined use of temporal and spatial information to achieve a highly sensitive detection process. Its sensitivity for benchmark videos was 98.5% and also had a low FN rate, which is vital for a fire detection system.

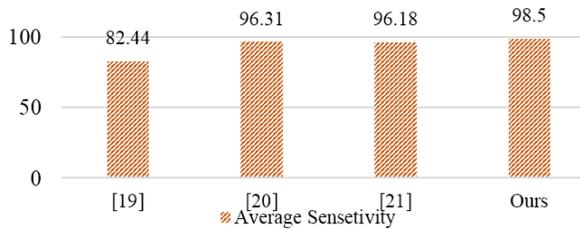

Fig 6. Comparison of flame detection performance for fire test videos using [19-21] and the proposed method.

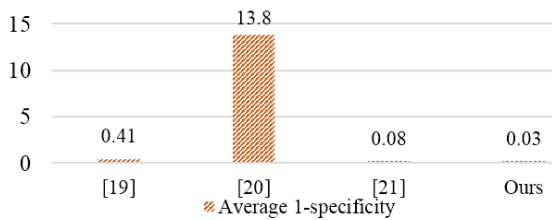

Fig 7. Comparison of flame detection performance for non-fire test videos using [19-21] and the proposed method.

Table II Comparison between proposed method, [2] and [22]

| methods | Video clips | | | | | | | |
|---|---|---|---|---|---|---|---|---|
| [22] | **100** | 94.29 | - | 96.15 | - | - | **95.20** | 100 |
| [2] | 95.70 | 100 | 100 | 93.10 | 100 | 100 | - | - |
| ours | 0.79 | **100** | **100** | **100** | **100** | **100** | 92 | 100 |

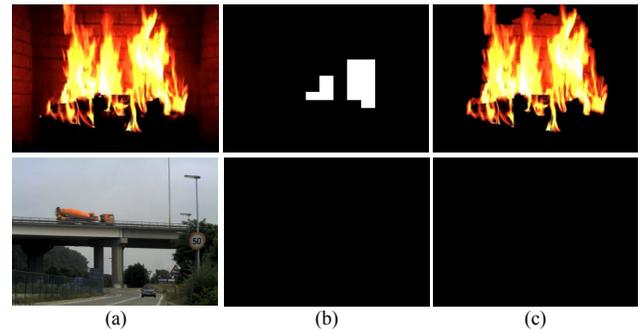

Figure 8. The output of the proposed method on fire and non-fire videos, (a) original image, (b) output of LBP-TOP, (c) output of our algorithm.